\documentclass[twocolumn,10pt]{asme2e}

\usepackage{epsfig} 
\usepackage{amsmath}
\confshortname{IDETC/CIE 2019}
\conffullname{the ASME 2019 \\International Design Engineering Technical Conferences \\
              and Computers and Information in Engineering Conference}

\confdate{18-21}
\confmonth{August}
\confyear{2019}
\confcity{Anaheim, CA}
\confcountry{USA}

\papernum{IDETC2019-97179}

\title{Topology Design and Position Analysis of a Reconfigurable Modular Hybrid-Parallel Manipulator}
\author{Rajashekhar Vachiravelu Saminathan 
    \affiliation{
	Researcher\\
	Tentacles Robotic Foundation\\
	Kanchipuram, Tamil Nadu, India 603202\\
    Email: vsrajashekhar@gmail.com
    }	
}

\begin{document}

\maketitle    

\begin{abstract}
{\it In the modern days, manipulators are found in the automated assembly lines of industries that produce products in masses. These manipulators can be used only in one configuration, that is either serial or parallel. In this paper, a new module which has two degrees of freedom is introduced. By connecting the two and three modules in series, 4 and 6 DoF hybrid manipulators can be formed respectively. By erecting 3 modules in parallel and with some minor modifications, a 6 DoF parallel manipulator can be formed. Hence the manipulator is reconfigurable and can be used as hybrid or parallel manipulator by disassembling and assembling. The topology design, forward and inverse position analysis has been done for the two hybrid configurations and the parallel configuration. This manipulator can be used in industries where flexible manufacturing system is to be deployed. The three configurations of the parallel manipulator has been experimentally demonstrated using a graphical user interface (GUI) control through a computer.           
}
\end{abstract}
\begin{nomenclature}
\entry{$\theta_i$}{Revolute joint variable of the $i^{th}$ joint}
\entry{$d_i$}{Prismatic joint variable of the $i^{th}$ joint}
\entry{$\alpha$}{The twist angle between the links}
\entry{$a$}{Length of the common normal}
\entry{$d$}{Length of the offset between z and to the common normal}
\entry{$SOC$}{Serial Open Chain}
\entry{$POC$}{Position and Orientation Characteristics}
\end{nomenclature}
\section{INTRODUCTION}
A modular reconfigurable manipulator is one that is made up of actuators, rigid links, passive joints and end effectors collectively. The modular reconfigurable manipulator shown in  \cite{chen2003interactive} has various one degree of freedom modules. These modules are either an active revolute or prismatic joint connected in series or parallel. This paper presents a mechanism module that has two degrees of freedom. It is actuated by two rotary actuators and no linear actuators. Although there are no active prismatic joints, there is a passive prismatic joint that plays a vital role in positioning the end effector when the two rotary actuators are activated.     

Among the manipulators available in the literature, only a few are reconfigurable in nature. This is due to minimal component level modularization \cite{huang2005conceptual}. This limitation is overcome in this paper by using a modified five-bar mechanism module. A five bar mechanism based manipulators that are found in the literature are being used as a laparoscopic manipulator \cite{kobayashi1999new} and a statically balanced parallel manipulator \cite{monsarrat2003workspace}.  

\begin{figure} 
\centerline{\psfig{figure=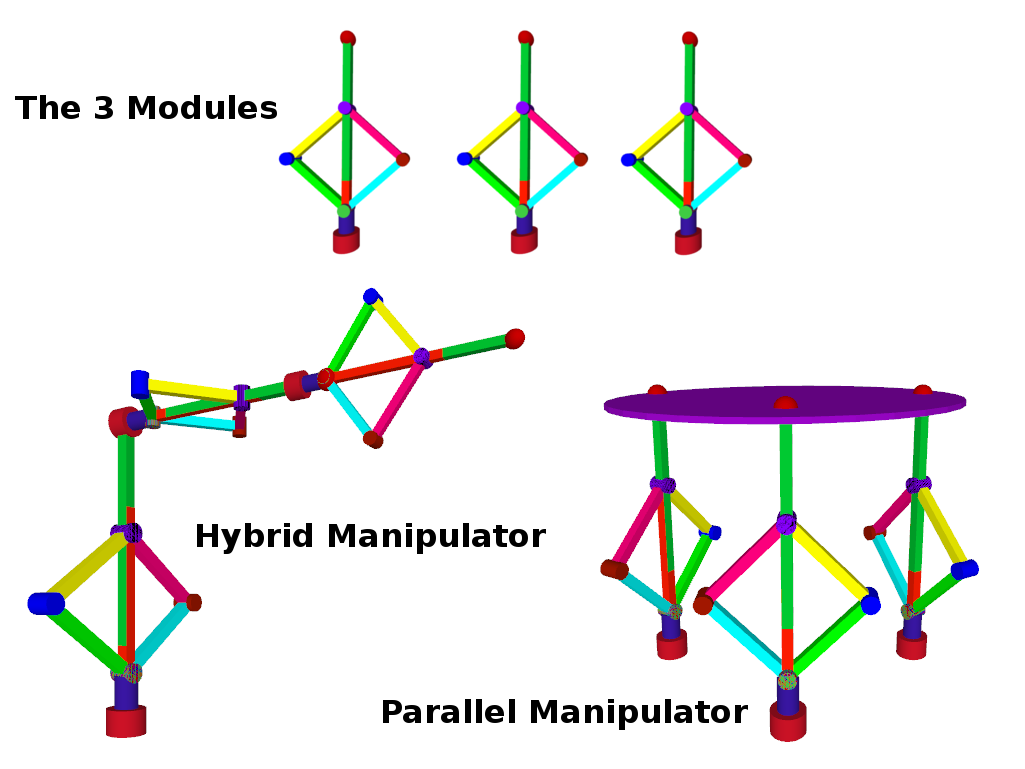,width=3.34in}}
\caption{The three modules reconfigurable into hybrid and parallel manipulator, both having 6 degrees of freedom}
\label{fig_first_three_modules.png}
\end{figure}

This paper mainly focuses on the forward and inverse kinematics of the three configurations of the modules namely 4 and 6 DoF hybrid manipulator, and 6 DoF parallel manipulator. It is as shown in Figure \ref{fig_first_three_modules.png}. The main contributions of the paper are as follows (1) A new modular mechanism has been presented that has two degrees of freedom. (2) The module is actuated only by two rotary actuators and no active prismatic joints. (3) When the three modules are connected in series or in parallel with some minor modifications, both the configurations produce a 6 degree of freedom manipulator. (4) A graphical user interface (GUI) for the manipulators has been created using the kinematic equations at the back-end in order to validate the proposed design.\\
  The paper is organized as follows: In section \ref{Sec:Module_Description}, the topology design, forward and inverse kinematic analysis of the module is done. In section \ref{sec:kinofhybmani} the DOF analysis, forward and inverse position analysis, and numerical investigation of the 4 and 6 DOF hybrid manipulator are done. In section \ref{sec:kinofparallelmani} the DOF analysis, forward and inverse position analysis of the parallel manipulator is done. The simulation and experimental results are shown in section \ref{sec:simu}. The possible applications of the manipulator are discussed in \ref{sec:appli}. The concluding remarks are given in section \ref{sec:conc}. 
\section{The RECONFIGURABLE MECHANISM MODULE}
\label{Sec:Module_Description}
\begin{figure} 
\centerline{\psfig{figure=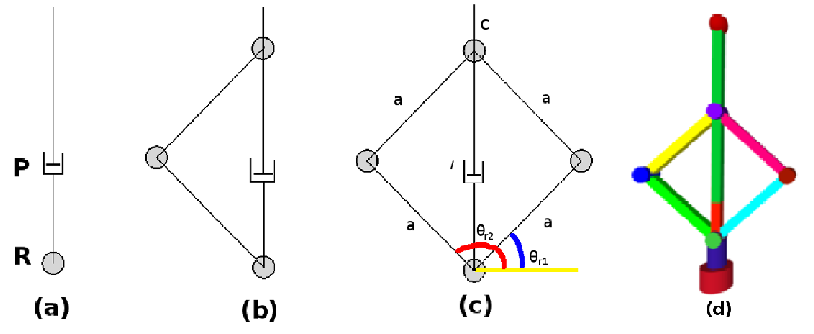,width=3.34in}}
\caption{Development of the 2 DoF module that forms an integral part of the hybrid and parallel manipulator. (a) The RP (Revolute-Prismatic) link. (b) The RRR (Revolute-Revolute-Revolute) jointed links attached to the previous link. (c) The final mechanism module that is proposed (d) The CAD model of the mechanism module that has two degrees of freedom}
\label{fig_module.png}
\end{figure}

The main mechanism module used in the manipulator is a five bar mechanism with a RP limb attached at the center. In other words, it can also be treated as two planar RR manipulators supporting a RP limb. There are two actuators attached one at the base of each RR manipulator. Upon actuation, the two actuators move the RR manipulators, which in turn moves the RP limb. It can be seen in the Figure \ref{fig_module.png}.
\subsection{Degrees of Freedom of the Module}
\label{subsec:dofofmodule}
The degrees of freedom of the module is calculated using the Grubler's mobility criterion \cite{norton1999design}. It is as follows: 
\begin{equation}
\label{equ_dof}
M=3L-2J-3G
\end{equation}
where,\\
M = Degrees of freedom\\
L = Number of links\\
J = Number of joints\\
G = Number of grounded links\\
In the module, there are 7 links, 8 joints and 1 grounded link. Substituting the values in equation \ref{equ_dof}, the degrees of freedom is obtained as 2.  \\

\begin{figure} 
\centerline{\psfig{figure=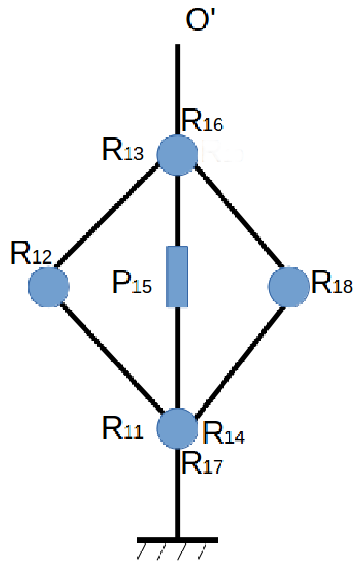,width=1.4in}}
\caption{Topology design of the mechanism module}
\label{fig_dof_module}
\end{figure}

The topology design is as shown in Figure \ref{fig_dof_module}. The degrees of freedom can also be obtained using the method given in \cite{yang2018topology}. It is as follows.
\begin{equation}
\label{equ_maindof1}
F= \sum_{i=1}^{m} f_i - \sum_{j=1}^{v} \xi_{Lj}
\end{equation}
\begin{equation}
\label{equ_maindof2}
\sum_{j=1}^{v} \xi_{Lj} = dim.{((\cap_{i=1}^{j} M_{b_{i}})\cup M_{b_{j+1}})}
\end{equation}
where,\\
$F$ - Degree of freedom of parallel manipulator(PM).\\
$f_{i}$ - Degree of freedom of the $i^{th}$ joint.\\
$m$ - Total number of joints in the parallel manipulator.\\
$v$ - Total number of independent loops in the mechanism, where $v=m-n+1$.\\
$n$ - Total number of links in the mechanism.\\
$\xi_{Lj}$ - Total number of independent equations of the $j^{th}$ loop.\\
$\cap_{i=1}^{j} M_{b_{i}}$ -  Position and orientation characteristic (POC) set generated by the sub-PM formed by the former j branches.\\
$M_{b_{j+1}}$ - POC set generated by the end link of j+1 sub-chains.\\

Calculating the number of independent loops putting $n=7$ and $m=8$, we get $v=2$.\\

The DOF of the parallel mechanism module is calculated based on the 8 steps as mentioned in \cite{yang2018topology} 
\begin{enumerate}
\item The topological structure of the mechanism is mentioned symbolically. \\
Branches:\\
$SOC_{1}({-R_{11}||R_{12}||R_{13}-})$\\
$SOC_{2}({-R_{14} \perp P_{15} \perp R_{16}-})$\\
$SOC_{3}({-R_{17}||R_{18}-})$\\
Platforms:\\
Fixed platform: $R_{11},R_{14},R_{17}$ are in same axis\\
Moving platform: $R_{13},R_{16}$ are in same axis parallel to the fixed platform\\
\item An arbitrary point $o'$ is chosen on the moving platform. \\
\item 
Determining the POC set of branches\\
\[M_{b_{1}}=\begin{pmatrix} t^2 (\perp R_{1i})\\
r^1 || (R_{1i})\\
\end{pmatrix}\quad\text{i=1,2,3} \]

\[M_{b_{2}}=\begin{pmatrix} t^2 (\perp R_{14}    ||P_{15} \perp R_{16})\\
r^1 (|| R_{14} || R_{16})\\
\end{pmatrix}\quad\text{} \]

\[M_{b_{3}}=\begin{pmatrix} t^2 (\perp R_{1k})\\
r^1 (|| R_{1k})\\
\end{pmatrix}\quad\text{k=7,8} \]
\item Finding the total number of independent displacement equations.\\  
Topological structure of the first independent loop\\
$\xi_{L_{1}}$ = dim($M_{b_{1}}$ $\cup$ $M_{b_{2}})$= \[dim\{\begin{bmatrix}
t^2 \\
r^1 (|| R_{11})\\ 
\end{bmatrix}
\cup
\begin{bmatrix}
t^2 \\
r^1 (|| R_{14})\\   
\end{bmatrix}\} 
\]
 \[=dim\{\begin{bmatrix}
t^2 \\
r^1 (||\diamond (R_{11},R_{14}))\\ 
\end{bmatrix}\}\quad\text{=3}
\]
Topological structure of the second independent loop\\
$\xi_{L_{2}}$ = dim(($M_{b_{1}}$ $\cap$ $M_{b_{2}}$) $\cup$ $M_{b_{3}})$ = \[dim\{(\begin{bmatrix}
t^2 \\
r^1 (|| R_{11})\\ 
\end{bmatrix}
\cap
\begin{bmatrix}
t^2 \\
r^1 (|| R_{14})\\   
\end{bmatrix}) 
\cup
\begin{bmatrix}
t^2 \\
r^1 (|| R_{17})\\   
\end{bmatrix}\}
\]

 \[=dim\{\begin{bmatrix}
t^2 \\
r^1 ( ||\diamond(R_{11} ||R_{14} ||R_{17}))\\ 
\end{bmatrix}\}\quad\text{=3}
\]
\item Calculating the DOF of the mechanism\\

$F= \sum_{i=1}^{m} f_i - \sum_{j=1}^{v} \xi_{Lj} = 8 - (3+3) = 2$
\item Finding the inactive pairs. \\
Based on the calculations done in \cite{yang2018topology}, similar steps were followed. It was found that without the limb $SOC_{2}({-R_{14} \perp P_{15} \perp R_{16}-})$, the same degrees of freedom were obtained when the revolute joints $R_{11}$, $R_{17}$ are activated. It is to be noted that the RP limb is used for easy positioning of the end effector, connecting two modules together and also plays a vital role in the forward and inverse kinematic analysis.
\item Determining the POC set of the parallel mechanism module.\\
Based on the formula given in \cite{yang2018topology},\\
$M_{Pa}=M_{b_{1}} \cap M_{b_{2}} \cap M_{b_{3}} $

\[M_{Pa}=\begin{bmatrix}
t^2 \\
r^1 (|| R_{11})\\ 
\end{bmatrix}
\cap
\begin{bmatrix}
t^2 (|| P_{15}) \\
r^1 (|| R_{14})\\   
\end{bmatrix} 
\cap
\begin{bmatrix}
t^2 \\
r^1 (|| R_{17})\\   
\end{bmatrix}
=
\begin{bmatrix}
t^1 \\
r^1 \\   
\end{bmatrix}
\]
The degree of freedom of the mechanism is 2 and the dimension of the above POC set is 2. Hence the module has one translational and one rotational degree of freedom. 
\item Select driving pairs\\
The joints $R_{1}$ and $R_{7}$ are chosen to be the driving pairs and hence they are fixed. Calculating the degrees of freedom of the new mechanism, \\
$F^{*}= \sum_{i=1}^{m} f_i - \sum_{j=1}^{v} \xi_{Lj} = 6- (3+3) = 0$\\
Since $F^{*}=0$, the joints $R_{1}$ and $R_{7}$ can be used as driving pairs simultaneously. 
\end{enumerate}

\subsection{Forward Kinematics of the Module}
\label{subsec:fwd_kin_of_module}
The forward kinematics can be done by following the steps below: \\
 \textbf{Step 1:} Get the $\theta_{r1}$ and $\theta_{r2}$ values as input.\\
 \textbf{Step 2:} Calculate the angle of rotation of the RP limb: \\ \begin{center}
$\theta$=($\theta_{r1}$+$\theta_{r2}$)/2 \\
\end{center}
 \textbf{Step 3:} Find the value of length of prismatic joint: \\
\begin{center}
$l$=2.a.$\cos(\alpha/2)$\\
$l$=2.a.$\cos((\theta_{r2}-\theta_{r1})/2)$\\
\end{center}
 \textbf{Step 4:} Add the extra length $c$ to the $l$ for the final positioning of the end effector:\\
\begin{center}
$l_t$=2.a.$\cos((\theta_{r2}-\theta_{r1})/2)+c$\\
\end{center}
 \textbf{Step 5:} Find the position of the end effector: 
\begin{center}
$x$=$l_{t}.\cos(\theta)$\\
$y$=$l_{t}.\sin(\theta)$
\end{center}

\subsection{Inverse Kinematics of the Module}
\label{subsec:inv_kin_of_module}
The inverse kinematics of the module is done as follows: \\
\textbf{Step 1:} Get the values of the $x$ and $y$ as input.\\ 
\textbf{Step 2:} Find the total length of the RP limb using the distance formula where $(x_{o},y_{o})$ is known :
  \begin{center}
    $l$=$\sqrt{(x_{o}-x)^2+(y_{o}-y)^2}$\\
    $l$=$l_t - c$\\
  \end{center}
\textbf{Step 3:} Find $(\theta_{r2}-\theta_{r1})$ \\
   \begin{equation}   
   \label{equ:twominusone}
   (\theta_r2-\theta_r1) = 2.\arccos(l,2a)\\
   \end{equation}
\textbf{Step 4:} Find $(\theta_{r1}+\theta_{r2})$ \\
    \begin{center}
    $\sin(\theta)$=$y/l_t$;\\
    $\cos(\theta)$=$x/l_t$ \\
    $\theta$ = $\arctan(\sin(\theta),\cos(\theta))$ \\
    \begin{equation}
    \label{equ:twoplusone}
    (\theta_{r1}+\theta_{r2})=2.\arctan(\sin(\theta),\cos(\theta))
    \end{equation}    
    \end{center}
\textbf{Step 5:} From equations \ref{equ:twominusone} and \ref{equ:twoplusone}, $\theta_{r1}$ and $\theta_{r2}$ can be found. 
\section{KINEMATICS OF THE HYBRID MANIPULATORS}
\label{sec:kinofhybmani}
Hybrid manipulators are those manipulators in which the configuration is a combination of both serial and parallel. A study of kinematics of hybrid manipulators can be found in \cite{tanev2000kinematics,shahinpoor1992kinematics, waldron1989kinematics}. The velocity analysis of the hybrid manipulators can be referred to in \cite{huang1993study}. In this section, the two configurations of hybrid manipulators designed are presented. The configuration 1 has four degrees of freedom and the configuration 2 has six degrees of freedom.  

\subsection{Configuration 1: Four DoF Hybrid Manipulator}
\begin{figure} 
\centerline{\psfig{figure=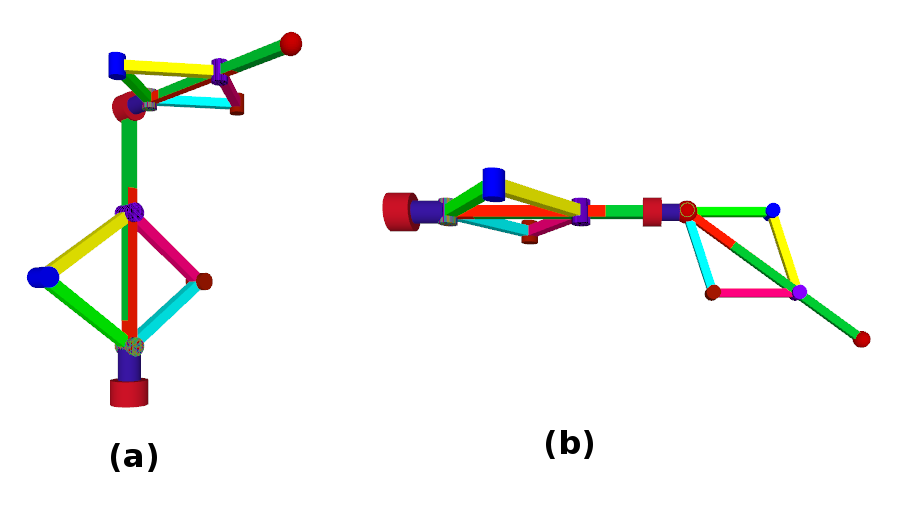,width=3.2in}}
\caption{The two modules connected in series forming the four degrees of freedom manipulator (a) Floor mounted (b) Wall mounted}
\label{fig_fourdoftwomodules.png}
\end{figure}

In this configuration, the manipulator has four degrees of freedom. There are two sub categories in the manipulator. In the first category as shown in Figure \ref{fig_fourdoftwomodules.png} (a), there are two modules of which the first module is mounted on the floor and the second module is attached to it in the perpendicular direction. In the second category as shown in Figure \ref{fig_fourdoftwomodules.png} (b), the manipulator is wall mounted.
  
\begin{figure} 
\centerline{\psfig{figure=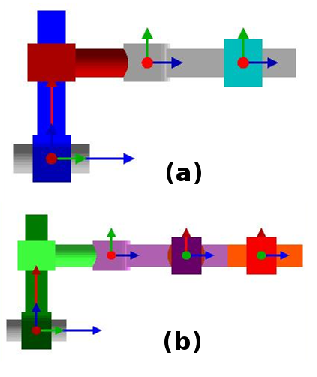,width=1.8in}}
\caption{The simplified model of the hybrid manipulator used for the forward and inverse position analysis (a) RPRP configuration (b) RPRPRP configuration}
\label{fig_simplifiedmodel.png}
\end{figure}
\subsubsection{Degrees of Freedom of the Configuration}
\begin{figure} 
\centerline{\psfig{figure=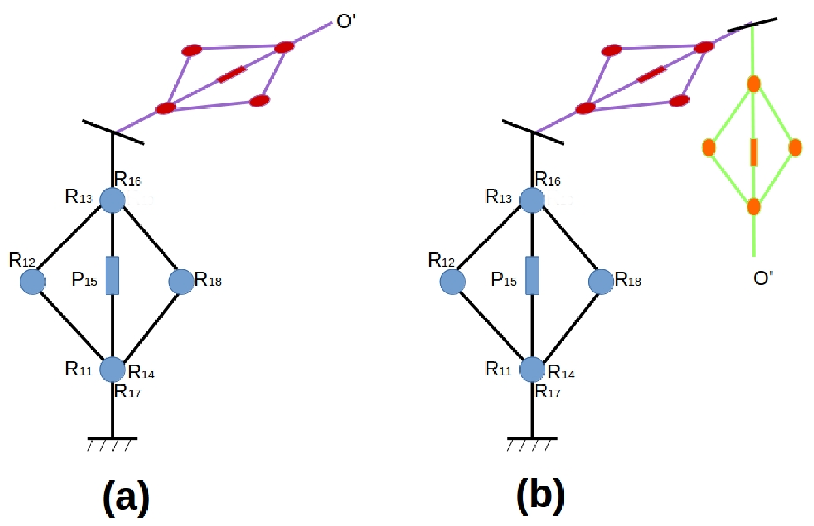,width=3.34in}}
\caption{Topology design of the hybrid manipulator (a) 4 DOF (b) 6 DOF}
\label{fig_dofhybrid}
\end{figure}
The topology design of the manipulator is as shown in Figure \ref{fig_dofhybrid} (a). The DOF of the parallel mechanism module is calculated based on the steps as mentioned in \cite{yang2018topology}.  
\begin{enumerate}
\item The topological structure of the mechanism is mentioned symbolically. \\
Branches:\\
$SOC_{i_1}({-R_{i1}||R_{i2}||R_{i3}-})$\\
$SOC_{i_2}({-R_{i4} \perp P_{i5} \perp R_{i6}-})$\\
$SOC_{i_3}({-R_{i7}||R_{i8}-})$\\
where i=1,2\\
Platforms:\\
Fixed platform: $R_{11},R_{14},R_{17}$ are in same axis\\
Moving platform: $R_{23},R_{26}$ are in same axis perpendicular to the fixed platform\\
\item An arbitrary point $o'$ is chosen on the moving platform. \\
\item Determining the POC set of branches is done as in step 3 of section \ref{subsec:dofofmodule}. Similarly it can be done for the other three branches. 
\item Finding the total number of independent displacement equations.\\ Dimension of the topological structure of the first and second independent loops in first module are calculated as in step 4 of section \ref{subsec:dofofmodule}. Similarly it can be done for the second module. 

\item Calculating the DOF of the mechanism\\

$F= \sum_{i=1}^{m} f_i - \sum_{j=1}^{v} \xi_{Lj} = 16 - (3+3+3+3) = 4$\\
Hence the manipulator has four degrees of freedom. The rest of the steps remain similar to those done in section \ref{subsec:dofofmodule}.
\end{enumerate}
\subsubsection{Forward Position Analysis}
\begin{table}[h]
\begin{center}
\caption{The DH parameters of the 4 DoF Hybrid Manipulator}
\label{table:4dofserial1}
\begin{tabular}{|c|c|c|c|c|}
\hline 
Joint Name & d  & $\theta$ & a  & $\alpha$ \\ 
\hline 
Revolute & 0 & $\theta_1$ & 0 & 90 \\ 
\hline 
Prismatic & $d_2$ & 90 & $L_2$ & 0 \\ 
\hline 
Revolute & 0 & $\theta_3$ & 0 & 90 \\ 
\hline 
Prismatic & $d_4$ & 0 & 0 & 0 \\ 
\hline 
\end{tabular}
\end{center} 
\end{table}
The forward position analysis is done as mentioned in section \cite{niku2001introduction}. The configuration is as shown in Figure \ref{fig_fourdoftwomodules.png} (a). The manipulator is simplified as shown in Figure \ref{fig_simplifiedmodel.png} (a) for the ease of kinematic analysis. The simplified version is a RPRP type serial manipulator. The DH parameters of the manipulator shown in Figure \ref{fig_simplifiedmodel.png} (a) is tabulated in Table \ref{table:4dofserial1}. A similar table can be formed for the manipulator shown in Figure \ref{fig_fourdoftwomodules.png} (b). The DH parameter values from the Table \ref{table:4dofserial1} are substituted in the transformation matrix $A_{i+1}^i$.  \\
\[A_{i+1}^i=\begin{pmatrix}\mathrm{cos}\left( \theta\right)  & -\mathrm{cos}\left( \alpha\right) \,\mathrm{sin}\left( \theta\right)  & \mathrm{sin}\left( \alpha\right) \,\mathrm{sin}\left( \theta\right)  & a\,\mathrm{cos}\left( \theta\right) \cr \mathrm{sin}\left( \theta\right)  & \mathrm{cos}\left( \alpha\right) \,\mathrm{cos}\left( \theta\right)  & -\mathrm{sin}\left( \alpha\right) \,\mathrm{cos}\left( \theta\right)  & a\,\mathrm{sin}\left( \theta\right) \cr 0 & \mathrm{sin}\left( \alpha\right)  & \mathrm{cos}\left( \alpha\right)  & d\cr 0 & 0 & 0 & 1\end{pmatrix}\]
The transformation matrix for the four linked RPRP serial manipulator is as follows.\\ 
\begin{equation}
\label{equ_4dof_fwd_kin}
A_4^0=A_1^0A_2^1A_3^2A_4^3
\end{equation}
where, 
\[A_{1}^0=\begin{bmatrix}\cos{\left( {{\theta}_{1}}\right) } & 0 & \sin{\left( {{\theta}_{1}}\right) } & 0\\
\sin{\left( {{\theta}_{1}}\right) } & 0 & -\cos{\left( {{\theta}_{1}}\right) } & 0\\
0 & 1 & 0 & 0\\
0 & 0 & 0 & 1\end{bmatrix}
;
A_{2}^1=\begin{bmatrix}0 & -1 & 0 & 0\\
1 & 0 & 0 & {{L}_{2}}\\
0 & 0 & 1 & {{d}_{2}}\\
0 & 0 & 0 & 1\end{bmatrix}
\]
\[A_{3}^2=\begin{bmatrix}\cos{\left( {{\theta}_{3}}\right) } & 0 & \sin{\left( {{\theta}_{3}}\right) } & 0\\
\sin{\left( {{\theta}_{3}}\right) } & 0 & -\cos{\left( {{\theta}_{3}}\right) } & 0\\
0 & 1 & 0 & 0\\
0 & 0 & 0 & 1\end{bmatrix}
;
A_{4}^3=\begin{bmatrix}1 & 0 & 0 & 0\\
0 & 1 & 0 & 0\\
0 & 0 & 1 & {{d}_{4}}\\
0 & 0 & 0 & 1\end{bmatrix}\]
\[A_{4}^0=\begin{pmatrix}n_x & o_x & a_x & p_x\\
n_y & o_y & a_y & p_y\\
n_z & o_z & a_z & p_z\\
0 & 0 & 0 & 1\end{pmatrix}\]
where,
$n_x=-\cos{\left( {{\theta}_{1}}\right) }\sin{\left( {{\theta}_{3}}\right) }$

$o_x=\sin{\left( {{\theta}_{1}}\right) }$

$a_x=\cos{\left( {{\theta}_{1}}\right) }\cos{\left( {{\theta}_{3}}\right) }$

$p_x={{d}_{4}}\cos{\left( {{\theta}_{1}}\right) }\cos{\left( {{\theta}_{3}}\right) }+{{d}_{2}}\sin{\left( {{\theta}_{1}}\right) }$

$n_y=-\sin{\left( {{\theta}_{1}}\right) }\sin{\left( {{\theta}_{3}}\right) }$

$o_y=-\cos{\left( {{\theta}_{1}}\right) } $

$a_y=\sin{\left( {{\theta}_{1}}\right) }\cos{\left( {{\theta}_{3}}\right) } $

$p_y={{d}_{4}}\sin{\left( {{\theta}_{1}}\right) }\cos{\left( {{\theta}_{3}}\right) }-{{d}_{2}}\cos{\left( {{\theta}_{1}}\right) }$

$n_z=\cos{\left( {{\theta}_{3}}\right) }$

$o_z=0$

$a_z=\sin{\left( {{\theta}_{3}}\right) }$

$p_z={{d}_{4}}\sin{\left( {{\theta}_{3}}\right) }+{{L}_{2}}$

\subsubsection{Inverse Position Analysis}
\label{subsubsec:inv_pos_ana}
In the inverse position analysis, the final position of the manipulator end effector is known and the joint variables have to be found out. It is done as explained in \cite{niku2001introduction}. From the equation \ref{equ_4dof_fwd_kin} and \ref{equ_4dof_inv_kin}, the values of $\theta_{1}$, $d_{2}$, $\theta_{3}$ and $d_{4}$ are obtained.  \\
\begin{equation}
\label{equ_4dof_inv_kin}
(A_1^0)^{-1}.A_4^0=A_2^1.A_3^2.A_4^3
\end{equation}
\[{{\theta}_{1}}=\operatorname{ARCTAN}\left( \frac{{{a}_{y}}}{{{a}_{x}}}\right) \]
\[{{\theta}_{3}}=\operatorname{ARCTAN}\left( \frac{{{a}_{z}}}{{{a}_{y}}\sin{\left( {{\theta}_{1}}\right) }+{{a}_{x}}\cos{\left( {{\theta}_{1}}\right) }}\right) \]
\[{{d}_{2}}={{p}_{x}}\sin{\left( {{\theta}_{1}}\right) }-{{p}_{y}}\cos{\left( {{\theta}_{1}}\right) }\]
\[{{d}_{4}}=\left( {{p}_{z}}-{{L}_{2}}\right) \sin{\left( {{\theta}_{3}}\right) }-\left( -{{p}_{y}}\sin{\left( {{\theta}_{1}}\right) }-{{p}_{x}}\cos{\left( {{\theta}_{1}}\right) }\right) \cos{\left( {{\theta}_{3}}\right) }\]
Substituting the values in the steps followed in section \ref{subsec:fwd_kin_of_module} and \ref{subsec:inv_kin_of_module}, the values of $\theta_{r1}$, $\theta_{r2}$, $\theta_{r3}$ and $\theta_{r4}$ are obtained. 
\subsubsection{Numerical Investigation of the Analysis}
(1) Forward Kinematics of the manipulator:\\
Let $\theta_{r1}=22$, $\theta_{r2}=23$, $\theta_{r3}=22$, $\theta_{r4}=23$, $a=100.0038078$, $L_{2}=50$. We get $\theta_{1}=22.5$,    $\theta_{3}=22.5$, $d_{2}=250$ and $d_{4}=250$. Substituting these values in equation \ref{equ_4dof_fwd_kin}, we get, \\
\[A_{4}^0=\begin{pmatrix}-0.3535 & 0.3827 & 0.8535 & 309.059\\
-0.1464 & -0.9239 & 0.3535 & -142.581\\
0.9239 & 0 & 0.3827 & 145.67\\
0 & 0 & 0 & 1\end{pmatrix}\]
(2) Inverse Kinematics of the manipulator:\\
(i) $\theta_{1}$ = $arctan(0.3535,0.8535)$ = 22.5\\
(ii) $\theta_{3}$ = $arctan(0.3827,0.9238)$ = 22.5\\ 
(iii) $d_{2}$ = 309.059 $\sin(22.5)$ +142.58 $\cos(22.5)$ = 250\\
(iv) $d_{4}$ = (145.67 - 50) $\sin(22.5)$ - (142.58 $\sin(22.5)$ -309.059 $\cos(22.5)$)$\cos(22.5)$ = 250\\
Following the steps in section \ref{subsec:fwd_kin_of_module} and \ref{subsec:inv_kin_of_module}, and substituting the above values, it can be easily shown that\\
$x=250\cos(22.5)=230.9699$\\
$y=250\sin(22.5)=95.6709$\\
From equation \ref{equ:twoplusone},\\
$\theta(r_{1})+\theta(r_{2})=2.arctan(95.6709,230.9699)$\\
$\theta(r_{1})+\theta(r_{2})=45$\\
From equation \ref{equ:twominusone},\\
$\theta(r_{2})-\theta(r_{1})=2.arccos(200,200.0076156)$\\
$\theta(r_{2})-\theta(r_{1})=1$\\
Solving for $\theta(r_{1})$ and $\theta(r_{2})$ we get, $\theta_{r1}=22$ and $\theta_{r2}=23$. This is found to be the equal to the input angles. In the similar way, it can be found that $\theta_{r3}=22$, $\theta_{r4}=23$.
\subsection{Configuration 2: Six DoF Hybrid Manipulator}
\begin{figure} 
\centerline{\psfig{figure=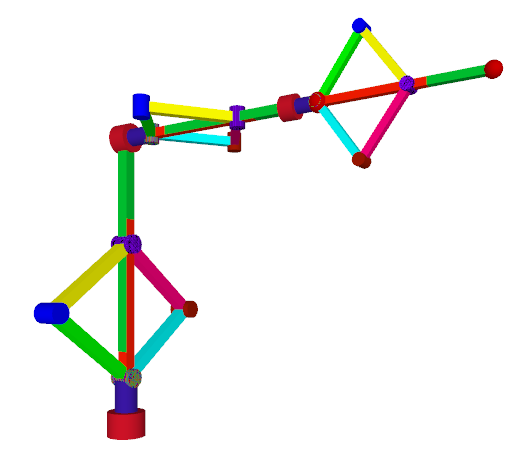,width=2.34in}}
\caption{The three modules connected in series forming the six degrees of freedom manipulator}
\label{fig_threemodulesinseries.png}
\end{figure}

Several hybrid manipulators that exist in the literature are listed and explained in \cite{gallardo2016kinematic}. In the hybrid manipulator designed in this paper, the three modules are connected in series. This is as shown in figure \ref{fig_threemodulesinseries.png}. 
\subsubsection{Degrees of Freedom of the Configuration}
The topology design of the manipulator is as shown in Figure \ref{fig_dofhybrid} (b). The DOF of the parallel mechanism module is calculated based on the steps mentioned in \cite{yang2018topology} 
\begin{enumerate}
\item The topological structure of the mechanism is mentioned symbolically. \\
Branches:\\
$SOC_{i_1}({-R_{i1}||R_{i2}||R_{i3}-})$\\
$SOC_{i_2}({-R_{i4} \perp P_{i5} \perp R_{i6}-})$\\
$SOC_{i_3}({-R_{i7}||R_{i8}-})$\\
where i=1,2,3\\
Platforms:\\
Fixed platform: $R_{11},R_{14},R_{17}$ are in same axis\\
Moving platform: $R_{33},R_{36}$ are in same axis perpendicular to the fixed platform\\
\item An arbitrary point $o'$ is chosen on the moving platform. \\
\item Determining the POC set of branches is done as in step 3 of section \ref{subsec:dofofmodule}. Similarly it can be done for the other six branches. 
\item Finding the total number of independent displacement equations.\\Dimension of the topological structure of the first and second independent loops in first module are calculated as in step 4 of section \ref{subsec:dofofmodule}. Similarly it can be done for the second and third module. 

\item Calculating the DOF of the mechanism\\

$F= \sum_{i=1}^{m} f_i - \sum_{j=1}^{v} \xi_{Lj} = 24 - (3+3+3+3+3+3) = 6$\\
Hence the manipulator has six degrees of freedom. The rest of the steps remain similar to those done in section \ref{subsec:dofofmodule}.
\end{enumerate}
\subsubsection{Forward Position Analysis}
\label{sec:fwdpositionanalysis}
The forward kinematics is done using the Denavit Hartenberg algorithm as mentioned in \cite{craig2005introduction}. The manipulator is simplified as shown in Figure \ref{fig_simplifiedmodel.png} (b) for the ease of kinematic analysis. The hybrid configuration is simplified as a R-P-R-P-R-P type serial manipulator for the ease of calculation. The Table \ref{table:6dofserial} shows the DH parameters for R-P-R-P-R-P type serial manipulator. The forward kinematics is calculated using the $A_{i+1}^i$ transformation matrix. \\
In each module, the value of $\theta_i$ of the revolute joint and $d_i$ of the prismatic joint are known. From these two values, the end effector value (of the module) can be found. \\
\begin{table}[h]
\begin{center}
\caption{The DH parameters of the 6 DoF Hybrid Manipulator}
\label{table:6dofserial}
\begin{tabular}{|c|c|c|c|c|}
\hline 
Joint Name & d  & $\theta$ & a  & $\alpha$ \\ 
\hline 
Revolute & 0 & $\theta_1$ & 0 & 90 \\ 
\hline 
Prismatic & $d_2$ & 90 & $L_2$ & 0 \\ 
\hline 
Revolute & 0 & $\theta_3$ & 0 & 90 \\ 
\hline 
Prismatic & $d_4$ & 90 & 0 & 90 \\ 
\hline 
Revolute & 0 & $\theta_5$ & 0 & -90 \\ 
\hline 
Prismatic & $d_6$ & 0 & 0 & 0 \\ 
\hline 
\end{tabular}
\end{center} 
\end{table}
%
%
%
%
%
The DH parameter values from the Table \ref{table:6dofserial} are substituted in transformation matrix $A_{i+1}^i$. There are six matrices for this configuration of the manipulator. They are multiplied together as shown in equation \ref{equ:homo6doftransformation}. The final matrix $A_6^0$ gives the homogeneous transformation matrix which carry the position and orientation information. Hence the forward position analysis is done.   
\begin{equation}
A_6^0=A_1^0A_2^1A_3^2A_4^3A_5^4A_6^5
\label{equ:homo6doftransformation}
\end{equation}
where, 
\[A_{1}^0=\begin{bmatrix}\cos{\left( {{\theta}_{1}}\right) } & 0 & \sin{\left( {{\theta}_{1}}\right) } & 0\\
\sin{\left( {{\theta}_{1}}\right) } & 0 & -\cos{\left( {{\theta}_{1}}\right) } & 0\\
0 & 1 & 0 & 0\\
0 & 0 & 0 & 1\end{bmatrix}
;
A_{2}^1=\begin{bmatrix}0 & -1 & 0 & 0\\
1 & 0 & 0 & {{L}_{2}}\\
0 & 0 & 1 & {{d}_{2}}\\
0 & 0 & 0 & 1\end{bmatrix}\]
\[A_{3}^2=\begin{bmatrix}\cos{\left( {{\theta}_{3}}\right) } & 0 & \sin{\left( {{\theta}_{3}}\right) } & 0\\
\sin{\left( {{\theta}_{3}}\right) } & 0 & -\cos{\left( {{\theta}_{3}}\right) } & 0\\
0 & 1 & 0 & 0\\
0 & 0 & 0 & 1\end{bmatrix}
;
A_{4}^3=\begin{bmatrix}0 & 0 & 1 & 0\\
1 & 0 & 0 & 0\\
0 & 1 & 0 & {{d}_{4}}\\
0 & 0 & 0 & 1\end{bmatrix}\]
\[A_{5}^4=\begin{bmatrix}\cos{\left( {{\theta}_{5}}\right) } & 0 & -\sin{\left( {{\theta}_{5}}\right) } & 0\\
\sin{\left( {{\theta}_{5}}\right) } & 0 & \cos{\left( {{\theta}_{5}}\right) } & 0\\
0 & -1 & 0 & 0\\
0 & 0 & 0 & 1\end{bmatrix}
;
A_{6}^5=\begin{bmatrix}1 & 0 & 0 & 0\\
0 & 1 & 0 & 0\\
0 & 0 & 1 & {{d}_{6}}\\
0 & 0 & 0 & 1\end{bmatrix}\]
\[A_{6}^0=\begin{pmatrix}n_x & o_x & a_x & p_x\\
n_y & o_y & a_y & p_y\\
n_z & o_z & a_z & p_z\\
0 & 0 & 0 & 1\end{pmatrix}\]

where,

$n_x=\cos{\left( {{\theta}_{1}}\right) }\cos{\left( {{\theta}_{3}}\right) }\sin{\left( {{\theta}_{5}}\right) }+\sin{\left( {{\theta}_{1}}\right) }\cos{\left( {{\theta}_{5}}\right) }$

$o_x=\cos{\left( {{\theta}_{1}}\right) }\sin{\left( {{\theta}_{3}}\right) }$

$a_x=\cos{\left( {{\theta}_{1}}\right) }\cos{\left( {{\theta}_{3}}\right) }\cos{\left( {{\theta}_{5}}\right) }-\sin{\left( {{\theta}_{1}}\right) }\sin{\left( {{\theta}_{5}}\right) }$

$p_x=\sin{\left({{\theta}_{1}}\right) }\left({{d}_{2}}-{{d}_{6}}\sin{\left({{\theta}_{5}}\right) }\right)+\cos{\left({{\theta}_{1}}\right)}\cos{\left( {{\theta}_{3}}\right) }\left( {{d}_{6}}\cos{\left( {{\theta}_{5}}\right) }+{{d}_{4}}\right)$

$n_y=\sin{\left( {{\theta}_{1}}\right) }\cos{\left( {{\theta}_{3}}\right) }\sin{\left( {{\theta}_{5}}\right) }-\cos{\left( {{\theta}_{1}}\right) }\cos{\left( {{\theta}_{5}}\right) } $

$o_y=\sin{\left( {{\theta}_{1}}\right) }\sin{\left( {{\theta}_{3}}\right) }$

$a_y=\cos{\left( {{\theta}_{1}}\right) }\sin{\left( {{\theta}_{5}}\right) }+\sin{\left( {{\theta}_{1}}\right) }\cos{\left( {{\theta}_{3}}\right) }\cos{\left( {{\theta}_{5}}\right) }$

$
p_y=\sin{\left( {{\theta}_{1}}\right) }\cos{\left( {{\theta}_{3}}\right) }\left( {{d}_{6}}\cos{\left( {{\theta}_{5}}\right) }+{{d}_{4}}\right)\\ -\cos{\left( {{\theta}_{1}}\right) }\left( {{d}_{2}}-{{d}_{6}}\sin{\left( {{\theta}_{5}}\right) }\right)$

$n_z=\sin{\left( {{\theta}_{3}}\right) }\sin{\left( {{\theta}_{5}}\right) }$

$o_z=-\cos{\left( {{\theta}_{3}}\right) } $

$a_z= \sin{\left( {{\theta}_{3}}\right) }\cos{\left( {{\theta}_{5}}\right) } $

$p_z= \sin{\left( {{\theta}_{3}}\right) }\left( {{d}_{6}}\cos{\left( {{\theta}_{5}}\right) }+{{d}_{4}}\right) +{{L}_{2}}$

\subsubsection{Inverse Position Analysis}
\label{sec:inversepositionanalysis}
The inverse kinematics can be found using the method mentioned in \cite{niku2001introduction} and done in section \ref{subsubsec:inv_pos_ana}. Using this method, the values of $\theta_i$ and $d_i$ can be found. 

$\theta_{1}=arctan(o_{y},o_{x})$

$\theta_{3}= arctan((-o_{y}\sin(\theta_{1})-o_{x}\cos(\theta_{1}),(o_{z}))$

$\theta_{5}=arctan((a_{y}\cos(\theta_{1})-a_{x}\sin(\theta_{1}),(n_{x}\sin(\theta_{1})-n_{y}\cos(\theta_{1})) $

The equation for $d_6$ is found to be coupled with $d_2$ and $d_4$. Hence as mentioned in \cite{merlet2006parallel} a sensor has to be placed at RP limb of the last module which directly gives the value of the $d_6$. The values of $d_2$ and $d_4$ can be found as follows.  

\[d_{2}=d_{6}\,\mathrm{\sin}( \theta_{5}) +p_x\mathrm{\sin}( \theta_{1}) -p_{y}\mathrm{\cos}( \theta_{1}) \]

\[d_{4}=\frac{p_{z}-L_{2}}{2\,\mathrm{\sin}\left( \theta_{3}\right) }-d_{6}\,\mathrm{\cos}\left( \theta_{5}\right) +\frac{p_{y}\,\mathrm{\sin}\left( \theta_{1}\right) +p_{z}\,\mathrm{\cos}\left( \theta_{1}\right) }{2\,\mathrm{\cos}\left( \theta_{3}\right) }\]

%
From these values, the actual value of the actuators $\theta_{ri}$ can be found as shown in section \ref{subsec:inv_kin_of_module}.    

\section{KINEMATICS OF THE PARALLEL MANIPULATOR}
\label{sec:kinofparallelmani}
Parallel manipulators consist of a fixed platform called the base and a moving platform at the top. In between these two platforms, there are a number of limbs that connect them together \cite{ceccarelli2013fundamentals}. 
In our case, the three modules (as explained in Section \ref{Sec:Module_Description}) are arranged parallel to each other forming a equilateral triangle at the base and the top.   
By using three legged parallel manipulators over six legged parallel manipulators, one can achieve better manipulability, increased workspace and lessen the presence of singularities in the manipulator \cite{gallardo2016kinematic}. \\

\subsection{Configuration of the Parallel Manipulator}
\begin{figure} 
\centerline{\psfig{figure=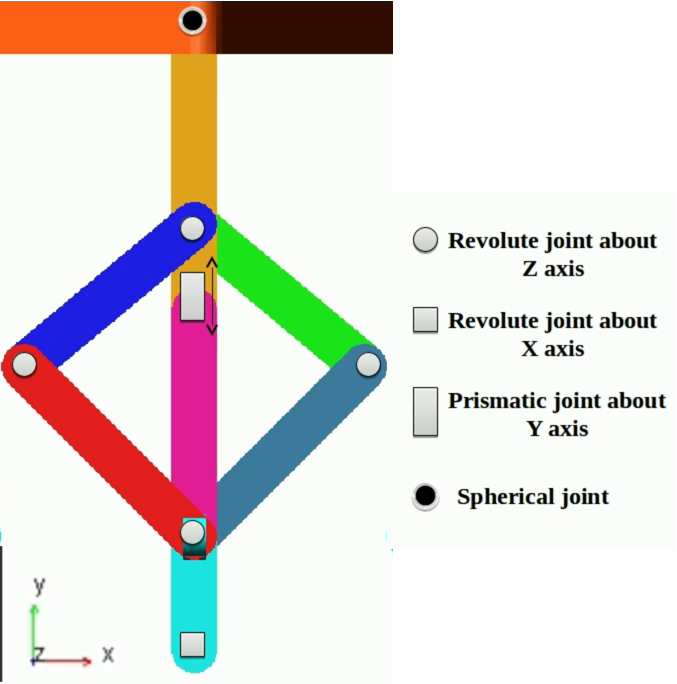,width=2.34in}}
\caption{The graphical representation of one limb of the parallel manipulator}
\label{fig_onelimbparallelmani.png}
\end{figure}
Among the parallel manipulators available in the literature, most of them are found to be using active prismatic joints. In the parallel manipulator designed in this paper, there are no active prismatic joints. There are three modules connected to the top plate (or the end effector plate) using spherical joints. The three modules are placed at an angle of 120 degree offset to each other for better stability and weight carrying ability. The bottom of these modules are connected to the revolute joints and then to the base. One of the limb used in the parallel manipulator is shown in Figure \ref{fig_onelimbparallelmani.png}. 

\subsection{Degrees of Freedom of the Parallel Manipulator}
\begin{figure} 
\centerline{\psfig{figure=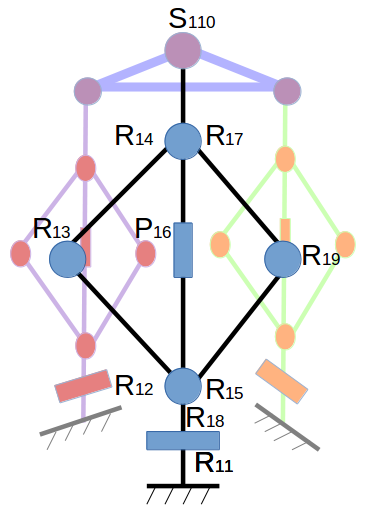,width=1.5in}}
\caption{Topology design of the parallel manipulator}
\label{fig_dofpm}
\end{figure}
The topology design of the parallel manipulator is as shown in Figure \ref{fig_dofpm}. The DOF of the parallel mechanism module is calculated based on the 8 steps as mentioned in \cite{yang2018topology}. 
\begin{enumerate}
\item The topological structure of the mechanism is mentioned symbolically. \\
Branches of the module:\\
$SOC_{i_1}({-R_{i2}||R_{i3}||R_{i4}-})$\\
$SOC_{i_2}({-R_{i5} \perp P_{i6} \perp R_{i7}-})$\\
$SOC_{i_3}({-R_{i8}||R_{i9}-})$\\
where i=1,2,3\\
Limbs of the parallel manipulator:\\
$SOC_{j_1}({-R_{j1}\perp(module)\perp S_{j11}-})$\\
where j=1,2,3\\
Platforms:\\
Fixed platform: $R_{11},R_{21},R_{31}$ are skew in space\\
Moving platform: $S_{110},S_{210},S_{310}$ are skew in space\\
\item An arbitrary point $o'$ is chosen on the moving platform. \\
\item Determining the POC set of branches\\
\[M_{b_{1}}=\begin{pmatrix} t^1 (\perp R_{11})\\
r^1 || (R_{11})\\
\end{pmatrix} \]

\[M_{b_{2}}=\begin{pmatrix} t^2 (\perp R_{1i})\\
r^1 || (R_{1i})\\
\end{pmatrix}\quad\text{i=2,3,4} \]

\[M_{b_{3}}=\begin{pmatrix} t^2 (\perp R_{14}    ||P_{15} \perp R_{16})\\
r^1 (|| R_{14} || R_{16})\\
\end{pmatrix}\quad\text{} \]

\[M_{b_{4}}=\begin{pmatrix} t^2 (\perp R_{1k})\\
r^1 (|| R_{1i})\\
\end{pmatrix}\quad\text{k=8,9} \]

\[M_{b_{5}}=\begin{pmatrix} t^3 (\perp R_{1l})\\
r^3 (|| R_{1l})\\
\end{pmatrix}\quad\text{l=10,11,12} \]
\item Finding the total number of independent displacement equations.\\
Topological structure of the first revolute joint\\
$Revolute$ $Joint$ = dim($M_{b_{1}}$)=2\\
Topological structure of the first independent loop\\
$\xi_{L_{1}}$ = dim($M_{b_{2}}$ $\cup$ $M_{b_{3}})$=3\\
Topological structure of the second independent loop\\
$\xi_{L_{2}}$ = dim(($M_{b_{2}}$ $\cap$ $M_{b_{3}}$) $\cup$ $M_{b_{4}})$ =3\\
Topological structure of the spherical joint\\
$Spherical$ $Joint$ = dim($M_{b_{5}}$)=2\\
Total number of independent displacement equations are \\
$dim(M_{serial}) = 2+3+3+2 = 10$

\item Calculating the DOF of the mechanism\\

$F= \sum_{i=1}^{m} f_i - \sum_{j=1}^{v} \xi_{Lj} = 36-(10+10+10)=6$\\
Hence the parallel manipulator has 6 DOF. The rest of the steps remain similar to those done in section \ref{subsec:dofofmodule}.
\end{enumerate}

\subsection{Forward Position Analysis}  
\begin{figure} 
\centerline{\psfig{figure=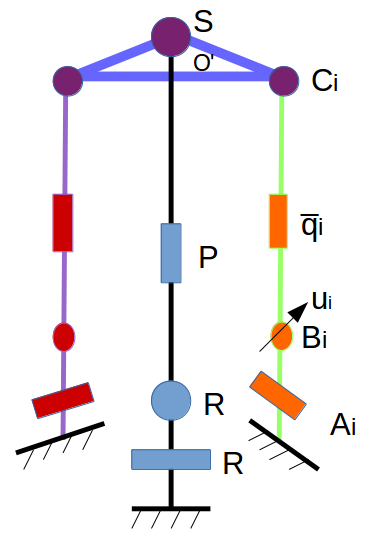,width=1.5in}}
\caption{The simplified 3-RRPS parallel manipulator.}
\label{fig_threerrpsmanipulator.png}
\end{figure}
The parallel manipulator designed is considered to be similar to the 3-RRPS configuration as shown in Figure \ref{fig_threerrpsmanipulator.png}. The RP link is considered to be the modular mechanism that is designed and presented in Section  \ref{Sec:Module_Description}. 

The equations needed for solving the forward kinematics of the 3-RRPS manipulator are written as follows as mentioned in \cite{gallardo2016kinematic}.

The spherical joints are constrained by the passive revolute joints by 
\begin{equation}
(c_i - b_i) \boldsymbol{\cdot} u_i=0
\label{equ:fwdparallel1}
\end{equation}
where, $u_i$ is a unit vector. \\

Based on the generalized coordinates $q_i$, the three loop closure equations are written as follows:
\begin{equation}
(c_i - b_i) \boldsymbol{\cdot} (c_i - b_i)= \bar{q}_i^2
\label{equ:fwdparallel2}
\end{equation}

The three compatibility equations are written from the equilateral triangle $\bigtriangleup$ $C_1, C_2, C_3$ as follows: 

\begin{equation}
(c_i - c_j) \boldsymbol{\cdot} (c_i - c_j)= e^2 \qquad    i, j = 1, 2, 3 mod(3).
\label{equ:fwdparallel3}
\end{equation}

By using Sylvester dialytic method of elimination, the equations \ref{equ:fwdparallel1}, \ref{equ:fwdparallel2}, \ref{equ:fwdparallel3} can be reduced into a 16th-order univariate polynomial equation. This leads the moving platform to at most 16 reachable poses. When the computation of the coordinates of spherical joints are done, the following homogeneous matrix is obtained.

\[T=\begin{pmatrix}R_m^0 & r_{o/O}\cr 0 & 1\end{pmatrix}\]

where, $R_m^0$ is the rotation matrix and $r_{o/O}$ is the position vector of the moving platform center.

\subsection{Inverse Position Analysis}

In the inverse position analysis, the position and orientation of the moving platform is known and the joint variables have to be calculated. 

The computation of the coordinates of points $C_i$ is computed as 
\begin{equation}
c_i=R_m^o d_i+r_{o/O}
\end{equation}

The value of $\bar{q}_i$ is computed from equation \ref{equ:fwdparallel2} as shown as follows:
\begin{equation}
\bar{q}_i= \sqrt{(c_i-b_i)\boldsymbol{\cdot}(c_i-b_i)}
\end{equation}

From equation \ref{equ:fwdparallel1}, the value of the $q_i$ can be computed as follows:
\begin{equation}
q_i=arctan(\frac{(c_i-b_i)\boldsymbol{\cdot}\hat{k}}{(c_i-b_i)\boldsymbol{\cdot}\hat{i}}
\end{equation}
From the values of $q_i$ and $\bar{q}_i$ the RP limb of the 3-RRPS manipulator is calculated. From these values, the two $\theta_{ri}$ values of the rotational joints can be calculated as done in Section \ref{subsec:inv_kin_of_module}. 
\section{SIMULATION AND EXPERIMENTS}
\label{sec:simu}
\begin{figure} 
\centerline{\psfig{figure=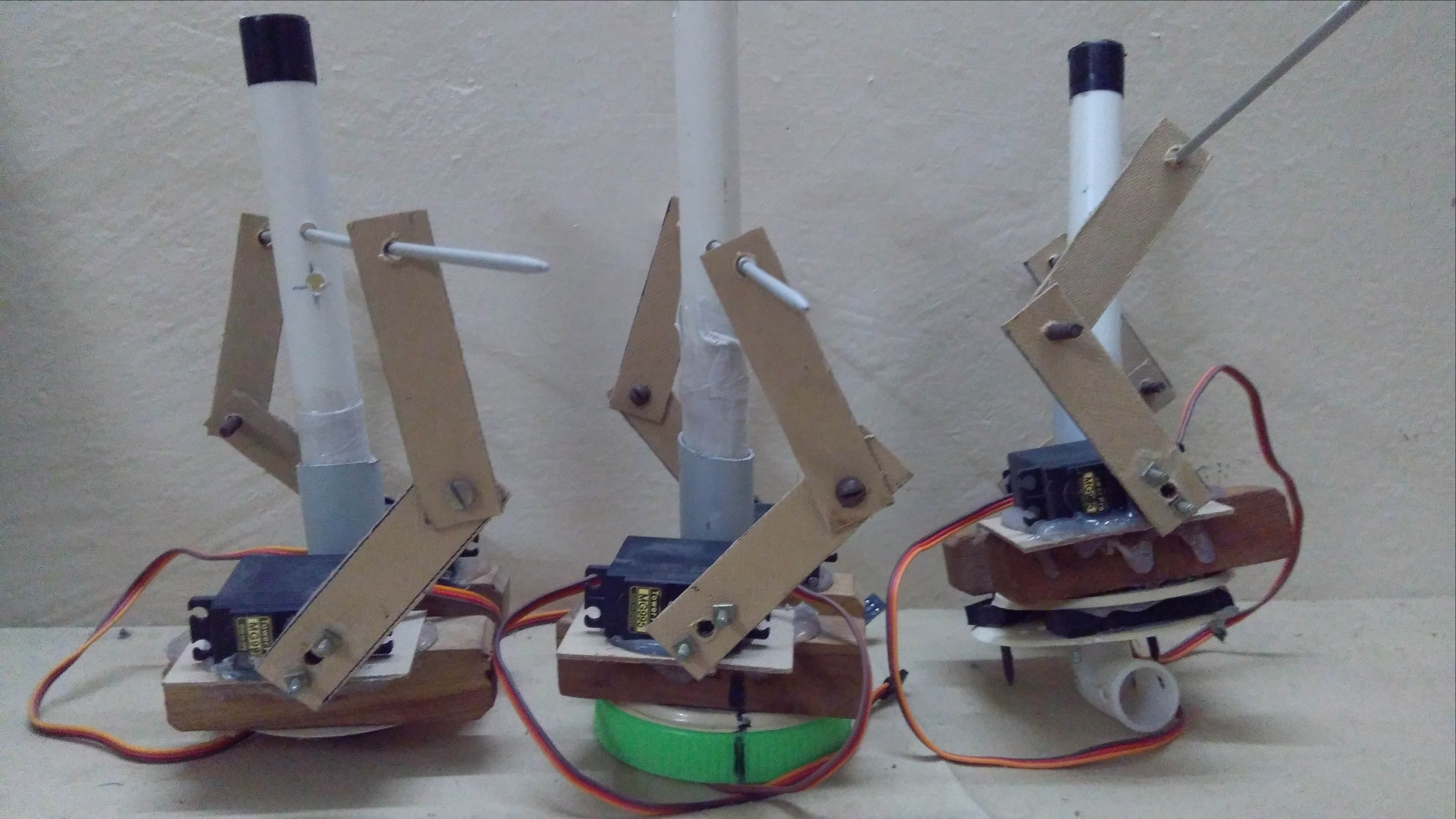,width=2.5in}}
\caption{The three modules used in the experiment to form the hybrid and parallel manipulator}
\label{fig_realmodules.png}
\end{figure}

\begin{figure} 
\centerline{\psfig{figure=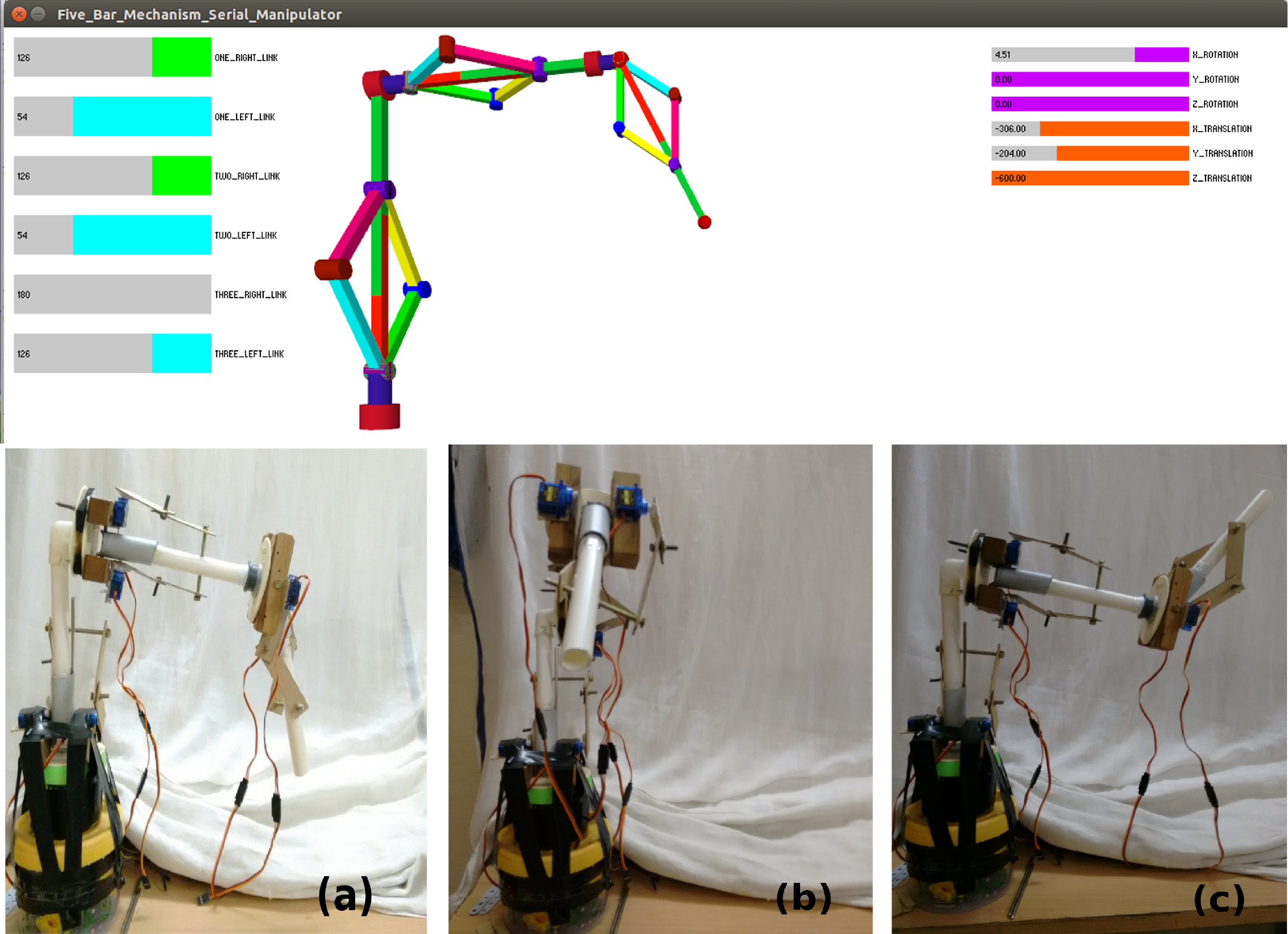,width=3.3in}}
\caption{Top: The GUI for the 6 DOF hybrid manipulator. Bottom: The three configurations of the hybrid manipulator during the experiment. The end effector (a) facing down (b) facing front (c) facing up }
\label{fig_hybridallinone.png}
\end{figure}

\begin{figure} 
\centerline{\psfig{figure=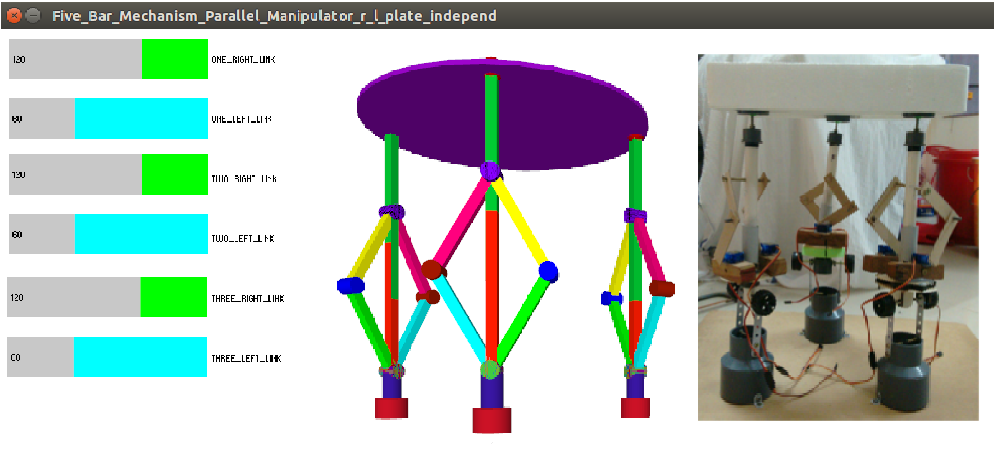,width=3in}}
\caption{The GUI of the six degrees of freedom parallel manipulator. The actual parallel manipulator is inserted on the right.}
\label{fig_real.png}
\end{figure}

The simulation was done using $Processing$ $IDE$ in $JAVA$ $ mode$. About 500 lines of code were written for each of the hybrid and parallel manipulator. The three modules were made in reality. The three modules used in the mechanism is as shown in Figure \ref{fig_realmodules.png}. The GUI and prototype of the hybrid and parallel manipulator are as shown in Figures \ref{fig_hybridallinone.png} and \ref{fig_real.png} respectively.  \\

\begin{figure} 
\centerline{\psfig{figure=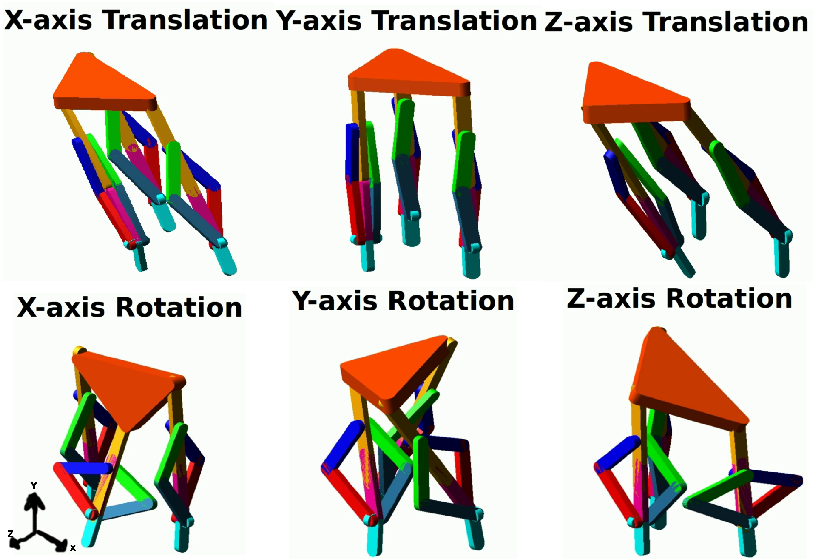,width=3.2in}}
\caption{The 6 degrees of freedom exhibited by the parallel manipulator}
\label{fig_sixdofparallel.png}
\end{figure}

The six degrees of freedom exhibited by the parallel manipulator is as shown in Figure \ref{fig_sixdofparallel.png}. The X-Y-Z translation and X-Y-Z rotation can be seen clearly in the figure. 

\section{APPLICATIONS OF RECONFIGURABLE MODULAR MANIPULATORS}
\label{sec:appli}
The mechanism module presented in section \ref{Sec:Module_Description} can be connected in series and also in parallel. The two scenarios where the reconfigurability and modularity can be utilized practically is as follows. When the modules are connected in series, they can be used in rehabilitation \cite{gan2016unified}. The proposed hybrid manipulator (connected in series) can be used as an arm or a leg of a human during rehabilitation as shown in Figure. The parallel manipulator can be used in wrist/ankle rehabilitation as previously mentioned in \cite{srivatsan2013position}.\\
An another scenario where both the hybrid and parallel manipulators can be used is in flexible manufacturing systems. The hybrid configuration can be used to do welding, painting, or pick and place operations when necessary. It can then be reconfigured to a parallel manipulator to perform machining operations.   
   

\section{CONCLUSIONS}
\label{sec:conc}
The topology design of the reconfigurable modular hybrid parallel manipulator was presented and the position analysis of it was done using the Denavit Hartenberg algorithm. It is to be noted that to the best of the author's knowledge, there is no manipulator mechanism that is actuated using only rotary actuators which is modular and reconfigurable, and can be used as a hybrid or parallel manipulator. The workspace analysis, singularity analysis, and velocity and acceleration analysis are considered to be done as future works. The authors have made this manipulator mechanism as a open source hardware and encourage the research community to develop it further.     
\bibliographystyle{asmems4}

\bibliography{asme2e}

%

\end{document}